\documentclass[sigconf]{acmart}

\renewcommand\footnotetextcopyrightpermission[1]{} 
\usepackage{cleveref}

\usepackage{bm}
\usepackage{bbm}
\usepackage{caption}
\usepackage{subcaption}
\usepackage{booktabs,amsfonts,dcolumn}
\usepackage{mathtools}
\usepackage{wasysym}
\usepackage{hhline}
\usepackage{float}
\usepackage{commath}

 \usepackage{multirow}
 \usepackage[ruled,vlined,linesnumbered]{algorithm2e}
\usepackage{booktabs,siunitx}

\usepackage{stmaryrd}
\usepackage{trimclip}
\usepackage{pifont}
\usepackage{enumitem}
\usepackage{mleftright}
\usepackage{makecell}
\usepackage{scalerel,stackengine}
\usepackage{bbold}
\setcopyright{acmcopyright}
\copyrightyear{2018}
\acmYear{2018}
\acmDOI{XXXXXXX.XXXXXXX}

\acmConference[Conference acronym 'XX]{Make sure to enter the correct
  conference title from your rights confirmation emai}{June 03--05,
  2018}{Woodstock, NY}
\acmPrice{15.00}
\acmISBN{978-1-4503-XXXX-X/18/06}

\begin{document}
\title{RbFT: Robust Fine-tuning for Retrieval-Augmented Generation against Retrieval Defects}

\author{Yiteng Tu}
\affiliation{%
    \institution{DCST, Tsinghua University}
    \city{Beijing}
    \country{China}
}
\email{yitengtu16@gmail.com}

\author{Weihang Su}
\affiliation{%
    \institution{DCST, Tsinghua University}
    \city{Beijing}
    \country{China}
}
\email{swh22@mails.tsinghua.edu.cn}

\author{Yujia Zhou}
\affiliation{%
    \institution{DCST, Tsinghua University}
    \city{Beijing}
    \country{China}
}

\author{Yiqun Liu}
\affiliation{%
    \institution{DCST, Tsinghua University}
    \city{Beijing}
    \country{China}
}

\author{Qingyao Ai\footnotemark}
\affiliation{%
    \institution{DCST, Tsinghua University}
    \city{Beijing}
    \country{China}
}
\email{aiqy@tsinghua.edu.cn}

\renewcommand{\shortauthors}{Yiteng Tu, Weihang Su, Yujia Zhou, Yiqun Liu, Qingyao Ai}

\begin{abstract}
Retrieval-augmented generation (RAG) enhances large language models (LLMs) by integrating external knowledge retrieved from a knowledge base. However, its effectiveness is fundamentally constrained by the reliability of both the retriever and the knowledge base. In real-world scenarios, imperfections in these components often lead to the retrieval of noisy, irrelevant, or misleading counterfactual information, ultimately undermining the trustworthiness of RAG systems.

To address this challenge, we propose Robust Fine-Tuning (RbFT), a method designed to enhance the resilience of LLMs against retrieval defects through two targeted fine-tuning tasks. Experimental results demonstrate that RbFT significantly improves the robustness of RAG systems across diverse retrieval conditions, surpassing existing methods while maintaining high inference efficiency and compatibility with other robustness techniques.
\end{abstract}



\maketitle

\section{Introduction} \label{sec:intro}
Large Language Models (LLMs) have achieved exceptional performance across diverse natural language processing tasks~\cite{mao2023gpteval,wang2022performance}, yet they remain constrained by challenges such as hallucinations, outdated or incomplete knowledge, and limited adaptability to specialized domains~\cite{izacard2023atlas,lewis2020retrieval}.
Retrieval-augmented generation (RAG) has emerged as a key technique to address these limitations by integrating LLMs with external knowledge sources, enabling enhanced factual accuracy, up-to-date information access, and improved domain-specific performance~\cite{li2024citation,huo2023retrieving,su2024unsupervised}.
Due to its effectiveness, RAG has been widely adopted to provide LLMs with a flexible mechanism for knowledge augmentation, enhancing their performance in varying scenarios~\cite{guu2020retrieval,chen2022murag}.

Despite their popularity, existing RAG systems suffer from a critical challenge: the performance of RAG systems heavily relies on the quality of the information provided by the retriever~\cite{yan2024corrective,li2022survey,tan2022tegtok}. 
Since the real-world retriever and its corresponding knowledge base could be defective and imperfect~\cite{shao2024scaling,dai2024unifying}, the retrieved documents provided to the LLM may contain inaccurate, irrelevant, or even malicious and misleading information~\cite{wang2024astute,chen2024benchmarking}.
Such low-quality or harmful information may hinder the LLM from accessing accurate knowledge, leading to inaccurate or misleading responses~\cite{xiang2024certifiably,zhang2023siren,yan2024corrective}, significantly degrading the performance and reliability of RAG systems (as discussed in \S\ref{subsec:ret_defects}).
Therefore, an imperative challenge arises: \textit{how to enhance the robustness of LLMs and RAG systems against retrieval defects, ensuring their ability to generate accurate responses even with defective retrieval results?}

Although there are several studies working on the robustness of RAG systems against retrieval defects through designing sophisticated inference mechanisms~\cite{xiang2024certifiably,yan2024corrective,wang2024astute,wei2024instructrag,weller2022defending}, they are limited in several perspectives. 
First, a critical issue is the significant increase in inference costs, which severely limits the RAG pipeline's runtime efficiency. 
This issue arises because existing studies mostly require generating intermediate results, which the LLM must aggregate or evaluate to produce the final response.
Second, more importantly, these methods tend to fail under severe retrieval defects or other challenging conditions since they do not fundamentally address the LLM's dependency on input knowledge.
For instance, RobustRAG~\cite{xiang2024certifiably} follows an "isolate-then-aggregate" pipeline, where answers are independently generated for each retrieved document and then aggregated. 
This approach not only incurs high inference costs but also becomes ineffective when the proportion of negative documents is high. 
CRAG~\cite{yan2024corrective}, on the other hand, leverages the large-scale web search to supplement and rely on the vanilla LLM to integrate and refine knowledge from different sources.
However, when the vanilla LLM fails to identify the defects in retrieved results, the whole pipeline would be broken and ineffective.

Based on the identified shortcomings of the aforementioned approaches, we believe that building an efficient and robust RAG system requires us to improve the inherent defensive capabilities of LLMs to fundamentally reduce their over-reliance on and blind trust in input information. 
Specifically, a well-defended LLM should have two characteristics: 
(1) \textbf{Defects Detection}: it should be capable of distinguishing what kind of information facilitates an effective response to the user's query and which documents are irrelevant or even harmful; 
(2) \textbf{Utility Extraction}: it should effectively utilize the limited useful information provided by the retriever while ignoring irrelevant or harmful content, even under adverse retrieval defects.
Therefore, we propose two corresponding fine-tuning tasks aimed at strengthening the LLM’s overall defensive capabilities: \textit{Defects Detection} and \textit{Utility Extraction}, collectively referred to as \textbf{R}o\textbf{b}ust \textbf{F}ine-\textbf{T}uning (RbFT). 
Specifically, we replace the original retrieved documents with defective ones and then train the LLM to (1) determine whether each document contains defects, thereby enhancing its ability to assess inputs critically; (2) generate the correct answers based on the defective inputs, improving its capability to utilize useful information effectively.
Experimental results demonstrate that our fine-tuning tasks can deliver superior performance in extremely challenging retrieval conditions, significantly outperforming the state-of-the-art baseline methods.

In summary, this paper makes three key contributions:
(1) We conduct a comprehensive analysis of potential retrieval defects in RAG systems from the perspectives of the retrieval system (i.e., the retriever and the corpus) and find that LLMs are highly vulnerable to retrieval defects. 
(2) We propose RbFT, a set of fine-tuning tasks that improve LLMs’ ability to evaluate and utilize retrieved information, enhancing robustness against retrieval defects.
(3) We conduct extensive experiments to show that RbFT can achieve superior performance under challenging retrieval conditions, significantly outperforming existing state-of-the-art methods. 

\section{Related Work} \label{sec:relat}
\subsection{Retrieval-Augmented Generation}

In recent years, Retrieval-Augmented Generation (RAG) has garnered widespread attention in the field of natural language processing (NLP) and shown significant advantages in knowledge-intensive tasks~\cite{borgeaud2022improving,lewis2020retrieval,wang2024knowledge,su2024stard,wang2024lekube,guu2020retrieval,izacard2020leveraging,dong2025decoupling,jiang2022retrieval}.
Traditional RAG typically follows the "Retrieval-then-Read" framework~\cite{borgeaud2022improving,guu2020retrieval,lewis2020retrieval,izacard2023atlas}, where an external retriever~\cite{zhai2008statistical,su2023caseformer,robertson2009probabilistic,su2023wikiformer,ma2023caseencoder} or a complex retrieval system~\cite{su2023thuir2,salemi2024towards} is adopted to search for relevant documents from a large-scale external corpus based on the user’s query. 
The retrieved documents provide external knowledge that supplements the query, allowing the generative model to incorporate relevant information beyond its parametric knowledge when generating a response.
To further enhance the retrieval effectiveness, researchers have introduced additional techniques such as query rewriting~\cite{ma2023query,dai2022promptagator} and re-ranking~\cite{arefeen2024leancontext,glass2022re2g} to refine the quality of retrieved documents before appending them to the generative model.

Building upon the traditional RAG framework, various extensions have been proposed to enhance its effectiveness and efficiency. 
One such extension, Parametric RAG~\cite{su2025parametric}, directly injects the retrieved documents into LLM parameters by offline parameterizing each document into independent plug-in parameters. During the inference process, the retrieved document’s parametric representation is merged and integrated into the LLM, enabling knowledge injection without extending the input context.
From another angle, GraphRAG~\cite{edge2024local,hu2024grag,peng2024graph} leverages pre-constructed knowledge graphs to retrieve graph elements with relational knowledge relevant to a given query. This approach has shown improved performance, especially in tasks that rely on structured and relational information.
Another research direction, dynamic RAG~\cite{jiang2022retrieval,su2024mitigating,su2024dragin}, dynamically triggers the retrieval module during the generation process when the LLM exhibits high uncertainty during the generation process.

In contrast to existing works that focus on improving retrieval quality, refining retrieval pipelines (e.g., through query rewriting or re-ranking), or reorganizing knowledge representation, we propose a fundamentally different perspective: directly enabling the LLM itself to handle imperfect or even malicious retrieval results.
Instead of assuming perfectly relevant or pre-filtered documents, we train the model to detect flaws in retrieved texts and extract only useful evidence, mitigating the impact of noisy, irrelevant, or incomplete information. This shift not only enhances accuracy but also fosters a more resilient and trustworthy RAG framework.

\begin{figure*}[t]
    \centering
    \includegraphics[width=0.98\textwidth]{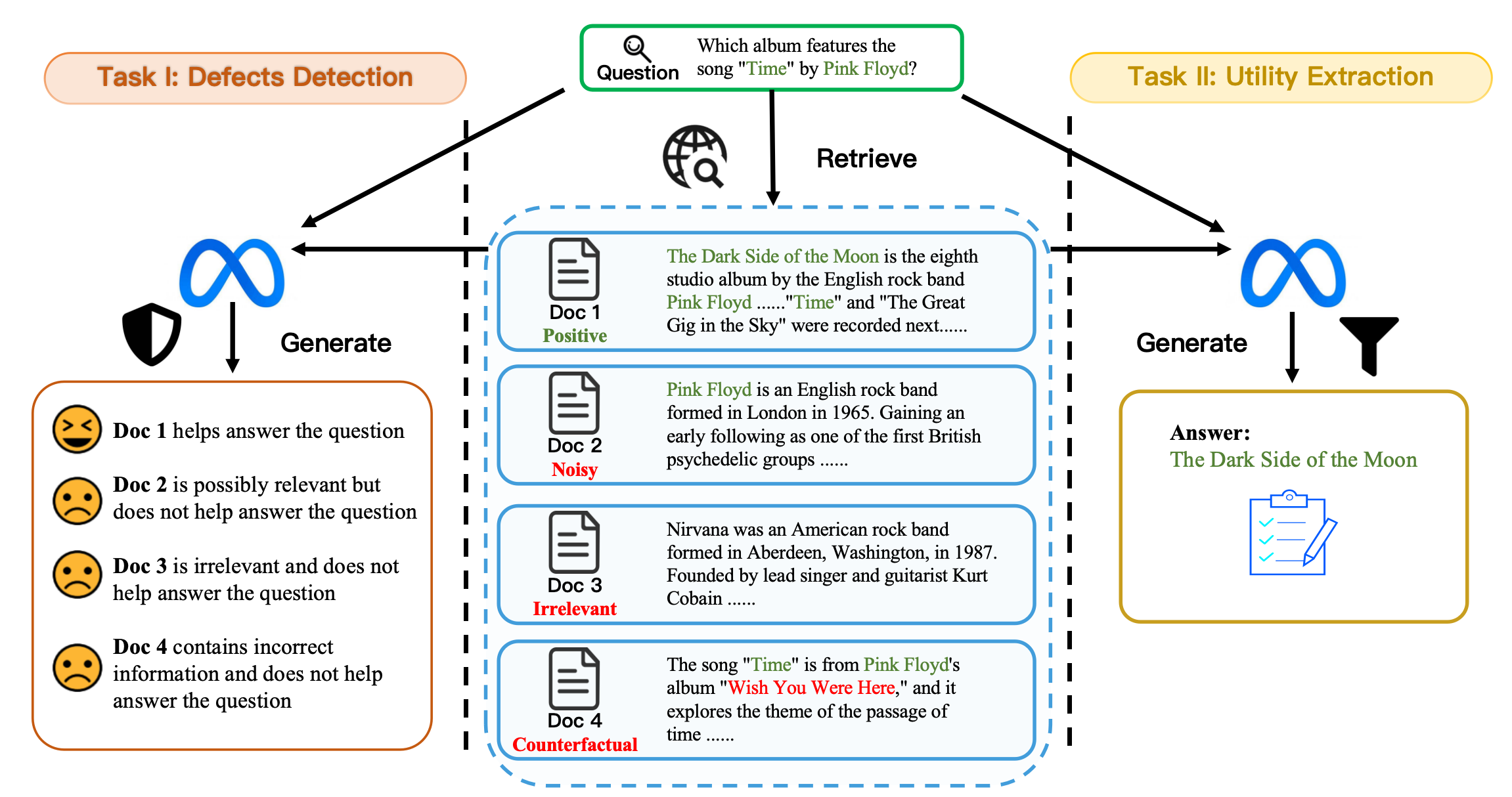}
    \caption{Overview of our RbFT. Specifically, RbFT consists of two sub-tasks: Defects Detection and Utility Extraction, which aim to identify the types of retrieval defects and generate the final answer with limited useful information, respectively. In the figure, \textcolor[rgb]{0.328,0.508,0.207}{green text} indicates relevant information, while \textcolor{red}{red text} represents incorrect counterfactual information.}
    \label{fig:method}
\end{figure*}

\subsection{Robustness in RAG}
The robustness of RAG systems refers to the ability of LLMs to consistently extract and apply relevant knowledge, even when exposed to varying or defective retrieval inputs~\cite{zhou2024trustworthiness}. 
Existing works have found that misinformation and corruption retrieval inputs pose significant challenges to the robustness of RAG systems.
Adversarial Addition and Modification~\cite{du2022synthetic} demonstrates the vulnerability of automated fact-checking systems when confronted with synthetic adversarial evidence. 
\citet{pan2021attacking} and \citet{pan2023risk} explore the threat posed by misinformation (whether manually crafted or generated by LLMs) to open-domain question-answering~(ODQA) systems, highlighting the vulnerability of these systems when exposed to misinformation corruption.
By injecting malicious texts into the knowledge base, PoisonedRAG~\cite{zou2024poisonedrag}, GARAG~\cite{cho2024typos} and Phantom~\cite{chaudhari2024phantom} can manipulate LLMs into generating specific incorrect or harmful responses.
To address these vulnerabilities, researchers have proposed strategies focusing on input optimization and knowledge integration.
~\citet{weller2022defending} conducts query augmentation and introduces a novel confidence method based on answer redundancy.
RobustRAG~\cite{xiang2024certifiably} employs an isolate-then-aggregate strategy to ensure the robustness of LLM responses against retrieval corruption attacks. 
By generating self-synthesized rationales, InstructRAG~\cite{wei2024instructrag} explicitly denoises the retrieved content, thereby enhancing the robustness of RAG systems.
CRAG~\cite{yan2024corrective} and AstuteRAG~\cite{wang2024astute} turn to refine and integrate knowledge derived from different sources to improve knowledge utilization and enhance the robustness of the generated answer.
However, despite the progress achieved by these methods, as discussed in \S\ref{sec:intro}, they can only partially control the retrieved content that the LLM accesses in RAG systems and fail to address the core issue of LLMs’ excessive reliance on the retrieval inputs. 
Unlike these works, we focus on enhancing the inherent defensive capabilities of LLMs to mitigate the impact of retrieval defects. 
By reducing dependency on external retrieval, our approach fundamentally improves RAG system robustness, offering a more resilient and trustworthy framework for real-world applications.

\section{Task Formulation} 
In this section, we first formalize the workflow of the RAG system and then discuss three types of potential retrieval defects that may occur in the RAG process.

\subsection{Workflow of RAG} \label{subsec:rag_workflow}
Following previous works~\cite{lewis2020retrieval,yan2024corrective}, a vanilla RAG system typically consists of a retrieval component $\mathcal{R}$, a generation component (i.e., the LLM) $\mathcal{G}$, and a corresponding corpus $\mathcal{C}=\{d\}$ containing a large collection of knowledge documents. 
Whenever the system receives a user query $q$, the retrieval component $\mathcal{R}$ first retrieves the top-$k$ most relevant documents $\mathcal{D}^q=\{d^q_1, d^q_2, ..., d^q_k\} \subset \mathcal{C}$ from the corpus $\mathcal{C}$:
\begin{equation}
    \mathcal{D}^q = \mathcal{R}(q,\ \mathcal{C})
\end{equation}
Then, the LLM $\mathcal{G}$ generates a response $r$ based on the query $q$ and relevant documents $\mathcal{D}^q$, where the expected output $r$ should ideally match the ground-truth answer $a$. Thus, the entire workflow can be formalized as:
\begin{equation}
    r = \mathcal{G}(q,\ \mathcal{D}^q) = \mathcal{G}(q,\ \mathcal{R}(q,\ \mathcal{C}))
\end{equation}
It is evident that, aside from the understanding and generation capabilities of the LLM itself, the quality of the generated response $r$ is highly dependent on the capability of the retriever $\mathcal{R}$ along with the quality of the corpus $\mathcal{C}$. 
Either an underperforming retriever or a low-quality corpus can significantly degrade the response quality. 
Since these retrieval-side issues are unavoidable in real-world scenarios, our work focuses on enhancing the capability of the generation component $\mathcal{G}$ to minimize their negative impact.

\subsection{Retrieval Defects} \label{subsec:ret_defects}
As mentioned above, due to the limitations of the retriever's performance and the quality of the corpus, the retrieval component often cannot guarantee that all returned documents fully meet the user query’s information needs, resulting in various types of retrieval defects. 
These defects can be broadly categorized into three types:
\subsubsection{Noisy Documents}
Noisy documents refer to content that is relevant to the query topic but does not directly answer the query. 
For example, given the query in Figure~\ref{fig:method}: "\textit{Which album features the song 'Time' by Pink Floyd?}", the retrieval system might return a general overview of the band \textit{Pink Floyd}~(Doc 2 in Figure~\ref{fig:method}). 
Although such a document is related to the band and its music, it does not explicitly mention the album containing the song "\textit{Time}", thereby failing to address the core question.

\subsubsection{Irrelevant Documents}
Irrelevant documents are those that bear no connection to the query topic. 
Such documents are typically retrieved due to inaccuracies in the retrieval model’s judgment.
For instance, in response to the same query, the system might retrieve a document introducing another band like \textit{Nirvana}~(Doc 3 in Figure~\ref{fig:method}). 
While about music, it has no relevance to \textit{Pink Floyd} or its albums, making it a clear example of an irrelevant document.

\subsubsection{Counterfactual Documents}
Counterfactual documents consist of false or misleading information, often resulting from inaccuracies or malicious manipulation within online content.
They fail to answer the question and may even lead to misconceptions.
For example, while the correct answer to the query is "\textit{The Dark Side of the Moon}", a counterfactual document might falsely claim that the song "\textit{Time}" appears on another album "\textit{Wish You Were Here}"~(Doc 4 in Figure~\ref{fig:method}).
Such incorrect information undermines the reliability of the retrieval process and can mislead both users and LLMs.

\
\begin{figure}[t]
    \centering
    \includegraphics[width=\columnwidth]{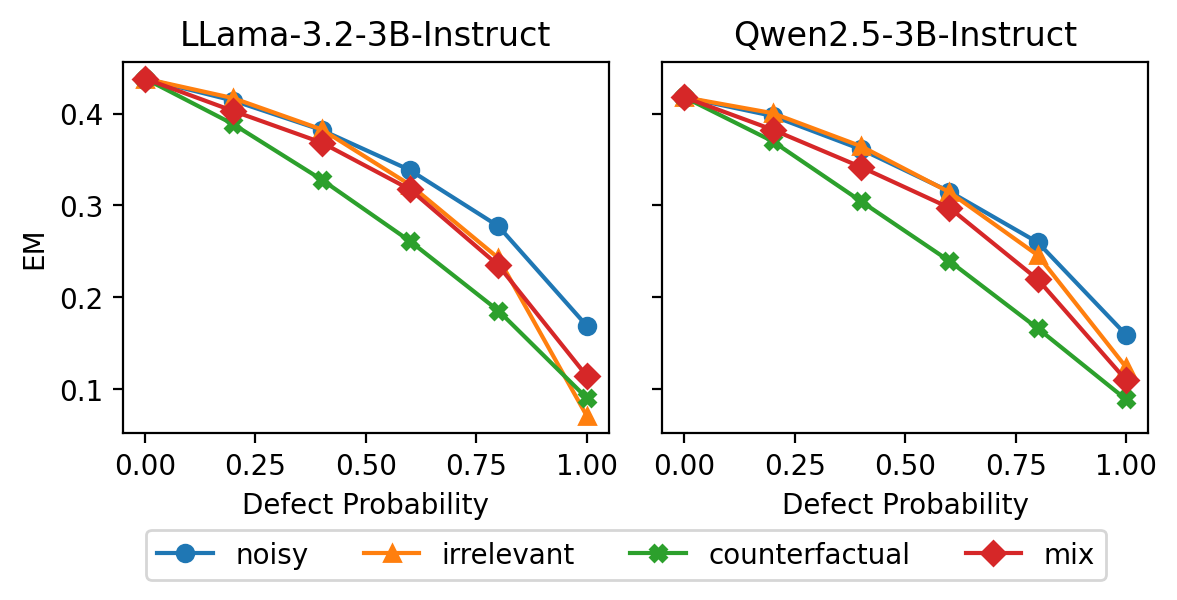}
    \caption{Emperical study: the impact of different types of retrieval defects on Vanilla RAG. The average EM metric on NQ, HQA, and TQA datasets is reported.}
    \label{fig:pre_exp}
\end{figure}

To simulate different levels of retrieval defects, we use varying probabilities $\tau$ to randomly replace the original retrieved documents with defective documents of different or identical types. 
Let the modified retrieval result be denoted as $\mathcal{D}^q_\tau$, our goal is for the LLM to generate the correct answer $a$ even when provided with $\mathcal{D}^q_\tau$.

To validate the negative impact of these retrieval defects on RAG systems, we conduct preliminary experiments on three datasets: Natural Questions~\cite{kwiatkowski2019natural}, HotpotQA~\cite{yang2018hotpotqa}, and TriviaQA~\cite{joshi2017triviaqa}, and report the average results. 
We select e5-base-v2~\cite{wang2022text} as the retriever, returning the top-5 retrieved results, while LLama-3.2-3B-Instruct~\cite{dubey2024llama} and Qwen2.5-3B-Instruct~\cite{yang2024qwen2} are adopted to generate the final answers, which are measured using the exact match (EM) metric. 
The tests are conducted using the aforementioned three types of retrieval defects and their mixture (i.e., randomly selecting one type of defective document for replacement). 
The defect replacement probability $\tau$ is set to \{0, 0.2, 0.4, 0.6, 0.8, 1.0\}. 
As observed in Figure~\ref{fig:pre_exp}, various types of retrieval defects pose a significant threat to the reliability of the vanilla RAG system.
Specifically, misleading counterfactual documents have the greatest impact on RAG systems, while even noisy documents that are relatively less harmful can be quite disruptive.
When all input documents are noisy, accuracy drops to below 40\% of the original.
Therefore, the severe impact of retrieval defects on the reliability of RAG systems underscores the urgent need to enhance their robustness.

\section{Robust Fine-Tuning~(RbFT)} \label{sec:method}
As mentioned in the \S\ref{sec:intro}, since the performance of an RAG system heavily depends on the quality of the retrieved documents and the model’s ability to effectively utilize them, we aim to enhance the robustness of the overall RAG system by fine-tuning the LLM to strengthen its intrinsic defensive capabilities. 
Here, the robustness is reflected not only in the model's ability to remain stable when confronted with retrieval defects or low-quality inputs but, more importantly, in its capacity to extract and utilize useful information effectively.
Specifically, we believe that a well-defended LLM should possess the following two key characteristics:

\begin{itemize}[leftmargin=*]

\item \textbf{Ability to assess the quality of input content.} 
A robust LLM should effectively distinguish the quality of documents, determining which documents are genuinely helpful in addressing the question and which are useless. 
On the one hand, distinguishing the input documents helps LLMs develop critical thinking and reduces their dependency on the retrieved results. 
On the other hand, accurate identification of useful information is also critical to prevent retrieval defects from affecting the final output.

\item \textbf{Ability to fully utilize useful information in the overall context.}
Once the model has developed an initial ability to assess input quality, it should be able to extract and exploit key information from high-quality and useful content while filtering out irrelevant or misleading content. 
Besides, the model should not only be capable of information filtering but also be able to synthesize multi-source information during answer generation to ensure the accuracy and completeness of the output.
\end{itemize}

Therefore, the core of our strategy lies in equipping LLMs with stronger self-detection and extraction capabilities, enabling them to maintain efficient and accurate outputs in complex real-world scenarios.
To achieve this goal, we design two specialized training tasks, namely \textit{Defect Detection} and \textit{Utility Extraction}, corresponding to input content assessment and effective information filtering, respectively~(as shown in Figure~\ref{fig:method}). 
The joint training of these tasks enables the LLM to improve its resistance to interference in complex input environments, thereby enhancing the overall robustness of the RAG system.

\subsection{Task I: Defects Detection}
The Defect Detection task aims to train the LLM to identify whether each retrieved document contributes to answering the user's query. 
If a document is useless, the LLM must also classify it into one of three defect types, i.e., noisy, irrelevant, or counterfactual document. 
We treat the original retrieved documents as positive examples and randomly replace them with different types of defective documents at a probability of $\tau$. 
To improve the efficiency, we adopt a listwise input format, where the LLM evaluates the entire list of retrieved documents at once. 
The prompt for this task is as follows:
\begin{center}
\fcolorbox{black}{gray!10}{\parbox{.99\linewidth}{
\textbf{Input}: \\
Determine whether the following documents help answer the given question. The assessment includes: \\
Assessment 1: The document helps answer the question. \\
Assessment 2: The document is possibly relevant but does not help answer the question. \\
Assessment 3: The document is irrelevant and does not help answer the question. \\
Assessment 4: The document contains incorrect information and does not help answer the question. \\
Only give me your assessment for each document and do not output any other words. \\ 
Documents: \\
Doc 1: \{\ document 1\ \} \\
Doc 2: \{\ document 2\ \} \\
...... \\
Question: \{\ question\ \} \\

\textbf{Output}: \\
Doc 1 helps answer the question. / Doc 1 is possibly relevant but does not help answer the question. / Doc 1 is irrelevant and does not help answer the question. / Doc 1 contains incorrect information and does not help answer the question. \\
Doc 2 ......
}}
\end{center}

\subsection{Task II: Utility Extraction}
In the Utility Extraction task, we aim to train the LLM to extract as much useful information as possible from the defective retrieval result. 
The LLM can either directly utilize the extracted relevant information or leverage the relevant context to activate its internal parametric knowledge to generate the correct answer. 
Meanwhile, Utility Extraction training also enables the LLM to directly and efficiently handle low-quality or contaminated contexts without prior cleanup.
Similarly, the original documents are replaced with defective documents at a probability $\tau$, and the LLM is required to answer based on these defective retrieval results while producing correct outputs. 
The prompt used for this task is as follows:

\begin{center}
\fcolorbox{black}{gray!10}{\parbox{.99\linewidth}{
\textbf{Input}: \\
Answer the question based on the given document. Only give me the answer and do not output any other words. \\
The following are given documents. \\
Doc 1: \{\ document 1\ \} \\
Doc 2: \{\ document 2\ \} \\
...... \\
Question: \{\ question\ \} \\

\textbf{Output}: \\
\{\ answer\ \}
}}
\end{center}

\subsection{Training Objective} 
RbFT aims to directly fine-tune the LLM using the two aforementioned tasks, thereby enhancing its organic defense capability. 
To achieve efficient training while preserving the LLM's general-purpose capabilities, we adopt the Low-Rank Adaptation (LoRA)~\cite{hu2021lora} technique for fine-tuning. 
Specifically, given the input text $x$, our goal is to maximize the probability of producing the correct output text $y$:
\begin{equation}
    \max_{\Theta} \sum_{(x, y)} \sum_{i=1}^{|y|} \log \left( p_{\Phi_0 + \Delta \Phi(\Theta)}(y_i | y_{< i}, x) \right)
\end{equation}
where $\Phi_0$ and $\Delta \Phi(\Theta)$ denote the LLM's original parameters and the learned parameter adjustments during fine-tuning, respectively.

\section{Experimental Settings} \label{sec:exp}
\subsection{Datasets and Evaluation Metrics}
We conduct experiments on three widely used Question Answering (QA) datasets: Natural Questions (NQ)~\cite{kwiatkowski2019natural}, HotpotQA (HQA)~\cite{yang2018hotpotqa}, and TriviaQA (TQA)~\cite{joshi2017triviaqa}, which cover both factoid QA and multi-hop QA tasks. 
Each data instance consists of a query and its corresponding ground truth answer. 
To create the training and validation sets, we randomly sample a total of 20,000 instances from the training splits of these three datasets, where 10\% of the training data is reserved for validation.
For evaluation, to ensure efficient experiments, following~\cite{wang2024astute,xiang2024certifiably}, we sample 1,000 queries from the test sets of NQ and TQA, as well as the HQA validation set (since HQA does not provide the test set), respectively (i.e., a total of 3,000 test queries).
The e5-base-v2~\cite{wang2022text} retriever is adopted to retrieve the top 100 most relevant documents for each query from the Wikipedia corpus~\footnote{\url{https://dl.fbaipublicfiles.com/dpr/wikipedia_split/psgs_w100.tsv.gz}}. 
To assess the performance of different RAG systems under varying levels of retrieval defects, we employ the standard QA evaluation metrics: exact match~(EM) and token-level F1 score~(F1), which measure the precision of the generated answers.

\subsection{Baselines}
Our RbFT is primarily compared with No RAG, Vanilla RAG, as well as four state-of-the-art robustness approaches for the RAG system: RobustRAG~\cite{xiang2024certifiably}, CRAG~\cite{yan2024corrective}, InstructRAG~\cite{wei2024instructrag} and AstuteRAG~\cite{wang2024astute}. 
RobustRAG leverages an "isolate-then-aggregate" strategy, where the LLM independently generates responses for each retrieved passage and then aggregates these individual responses to produce the final output. 
CRAG introduces a lightweight retrieval evaluator that triggers different knowledge retrieval actions based on evaluation results and enables knowledge refinement. 
To ensure a fair comparison of different RAG systems, we disable the module in CRAG that is responsible for large-scale web searches to acquire additional knowledge. 
InstructRAG instructs LLMs to explicitly denoise retrieved content by generating self-synthesized explanatory rationales, which explain how the answer is derived from the retrieved documents.
AstuteRAG, on the other hand, focuses on resolving conflicts between the internal knowledge of the LLM and the external knowledge provided by the retriever. 
It achieves this goal through an "iterative source-aware knowledge consolidation" process that integrates the two kinds of knowledge and handles knowledge conflicts.

\subsection{Data Generation}
We simulate three types of defective documents using different approaches. 
For noisy and irrelevant documents, inspired by the negative sampling methods commonly used in training dense retrieval models~\cite{karpukhin2020dense, qu2020rocketqa,zhan2021optimizing}, these two types of defective documents can be analogized to hard negatives and random negatives, respectively. 
Accordingly, noisy documents can be obtained by randomly sampling from lower-ranked retrieval results (e.g., documents ranked after 50 in the retrieval results), while irrelevant documents can be randomly sampled from the entire corpus.
For counterfactual documents, we adopt a two-step generation strategy: first, given the query, the correct answer, and the original retrieval results, we use Llama-3.2-3B-Instruct~\cite{dubey2024llama} to generate a misleading incorrect answer. 
Then, we call the LLM again to rewrite all original documents by replacing all information related to the correct answer with the misleading incorrect answer:
\begin{center}
\fcolorbox{black}{gray!10}{\parbox{.99\linewidth}{
\textbf{Step 1 Input}: \\
Based on a given question and its correct answer, generate a misleading wrong answer. You can refer to some relevant documents for inspiration. The wrong answer should belong to the same entity type as the correct answer (e.g., person, time, place, organization, data, etc.) to enhance its confusion. If the answer does not contain an entity, replace a key entity in the question and treat it as the wrong answer.  Only give me the wrong answer and do not output any other words.\\
The following are given documents. \\
Doc 1: \{\ document 1\ \} \\
Doc 2: \{\ document 2\ \} \\
...... \\
Question: \{\ question\ \} \\
Correct Answer: \{\ answer\ \}\\

\textbf{Step 2 Input}: \\
You are a writing AI. Rewrite the passage by replacing all content and information related to \{\ correct answer\ \} with \{\ wrong answer\ \}. Ensure that the rewritten passage is fluent and concise, maintaining a language style similar to the original. Only give me the rewritten passage and do not output any other words. \\
Original Document: \{\ document\ \}
}}
\end{center}

\begin{table*}
  \caption{
The average evaluation results of each model on the three datasets under the Clean~($\tau=0$), Normal~($\tau=0.4$), and Hard~($\tau=1.0$) settings. "*" refers to a significant improvement compared to the Vanilla RAG baseline at $p < 0.05$ level using the two-tailed pairwise t-test. The best and second-best methods are marked in bold and underlined, respectively. The improvement ratio of the best model over the second-best model is also reported.}
  \label{tab:mainresult}
  \renewcommand{\arraystretch}{1.3}
  \scalebox{0.75}{
\begin{tabular}{c|c||cc||cc|cc|cc|cc||cc|cc|cc|cc}
\toprule
\multirow{3}{*}{LLM} & \multirow{3}{*}{Method} & \multicolumn{2}{c||}{\multirow{2}{*}{Clean $(\tau=0)$}} & \multicolumn{8}{c||}{Normal $(\tau=0.4)$} & \multicolumn{8}{c}{Hard $(\tau=1.0)$} \\ \cline{5-20} 
 &  & & & \multicolumn{2}{c|}{Noisy} & \multicolumn{2}{c|}{Irrelevant} & \multicolumn{2}{c|}{Counterfactual} & \multicolumn{2}{c||}{Mix} & \multicolumn{2}{c|}{Noisy} & \multicolumn{2}{c|}{Irrelevant} & \multicolumn{2}{c|}{Counterfactual} & \multicolumn{2}{c}{Mix} \\
 \cline{3-20}
 &  & EM & F1 & EM & F1 & EM & F1 & EM & F1 & EM & F1 & EM & F1 & EM & F1 & EM & F1 & EM & F1 \\ 
 \toprule
\multirow{9}{*}{Llama} 
 & No RAG & 26.5 & 34.7 & - & - & - & - & - & - & - & - & - & - & - & - & - & - & - & - \\
 & Vanilla RAG & \underline{43.8} & \underline{52.7} & \underline{38.2} & \underline{46.2} & \underline{38.3} & \underline{46.3} & \underline{32.8} & 40.9 & \underline{36.8} & \underline{45.2} & 16.9 & 22.0 & 7.1 & 9.5 & 9.0 & 15.4 & 11.4 & 17.1 \\
 \cline{2-20}
 & RobustRAG  & 32.1 & 43.5 & 26.7 & 37.0 & 27.2 & 37.6 & 24.9 & 35.2 & 26.9 & 37.5 & 15.0 & 22.2 & 4.0 & 5.8 & 10.1 & 18.4* & 12.6 & 19.7* \\
 & CRAG  & 39.4 & 48.5  & 35.0 & 43.1 & 35.7 & 43.6 & 30.4 & 39.3 & 32.9 & 41.6 & 21.0* & 27.4* & 21.9* & 27.6* & 13.5* & 20.7* & 17.3* & 23.8* \\
 & InstructRAG  & 38.0 & 47.6 & 32.4 & 41.3 & 32.6 & 41.3 & 27.7 & 36.6 & 31.4 & 40.2 & 14.7 & 21.3 & 4.0 & 8.3 & 9.6 & 16.6* & 11.2 & 17.2 \\
 & AstuteRAG  & 37.6 & 47.5 & 34.4 & 43.8 & 34.6 & 43.7 & 32.2 & \underline{41.5} & 33.2 & 42.7 & \underline{23.8}* & \underline{32.1}* & \underline{22.1}* & \underline{30.2}* & \underline{19.6}* & \underline{28.4}* & \underline{21.1}* & \underline{29.6}* \\
 & RbFT (Ours) & \textbf{48.4}* & \textbf{58.5}* & \textbf{44.5}* & \textbf{53.9}* & \textbf{44.3}* & \textbf{53.7}* & \textbf{44.2}* & \textbf{54.2}* & \textbf{44.4}* & \textbf{54.0}* & \textbf{31.1}* & \textbf{39.1}* & \textbf{28.2}* & \textbf{36.4}* & \textbf{33.8}* & \textbf{43.1}* & \textbf{31.9}* & \textbf{40.9}* \\ 
 \cline{2-20}
 & Improvement & 10.5\% & 11.0\% & 16.5\% & 16.7\% & 15.7\% & 16.0\% & 37.3\% & 29.4\% & 20.7\% & 19.5\% & 30.7\% & 21.8\% & 27.6\% & 20.5\% & 72.4\% & 51.8\% & 51.2\% & 38.2\%\\
 \bottomrule
\multirow{9}{*}{Qwen} 
 & No RAG & 20.8 & 27.5 & - & - & - & - & - & - & - & - & - & - & - & - & - & - & - & - \\
 & Vanilla RAG  & \underline{41.8} & \underline{50.7} & \underline{36.1} & \underline{44.3} & \underline{36.4} & \underline{44.6} & \underline{30.4} & 38.6 & \underline{34.2} & \underline{42.6} & 15.9 & 21.6 & 12.4 & 15.5 & 8.9 & 15.4 & 11.0 & 16.9 \\
 \cline{2-20}
 & RobustRAG  & 25.8 & 37.6 & 22.5 & 33.0 & 22.5 & 33.2 & 19.9 & 29.9 & 20.3 & 31.3 & 9.9 & 15.7 & 2.0 & 3.3 & 7.7 & 14.9 & 8.4 & 14.7 \\
 & CRAG  & 37.0 & 45.5 & 31.1 & 39.0 & 32.5 & 40.0 & 27.5 & 35.3 & 30.2 & 38.4 & 17.0* & 22.6 & 13.9* & 17.2* & 11.4* & 17.6* & 12.4* & 18.0 \\
 & InstructRAG  & 35.4 & 46.6 & 30.6 & 40.5 & 30.8 & 40.6 & 25.9 & 35.9 & 29.1 & 38.9 & 14.8 & 20.9 & 8.7 & 13.1 & 8.1 & 15.5 & 10.1 & 16.6 \\
 & AstuteRAG  & 35.9 & 46.3 & 32.3 & 41.9 & 32.3 & 41.6 & 29.8 & \underline{39.2} & 30.8 & 40.4 & \underline{18.6}* & \underline{26.0}* & \underline{15.7}* & \underline{21.8}* & \underline{15.3}* & \underline{23.4}* & \underline{16.4}* & \underline{24.1}* \\
 & RbFT (Ours) & \textbf{45.4}* & \textbf{56.8}* & \textbf{40.8}* & \textbf{51.7}* & \textbf{41.0}* & \textbf{51.6}* & \textbf{40.4}* & \textbf{51.5}* & \textbf{40.6}* & \textbf{51.5}* & \textbf{24.6}* & \textbf{33.3}* & \textbf{21.4}* & \textbf{29.9}* & \textbf{25.1}* & \textbf{34.7}* & \textbf{24.0}* & \textbf{33.2}* \\
 \cline{2-20}
 & Improvement & 8.6\% & 12.0\% & 13.0\% & 16.7\% & 12.6\% & 15.7\% & 32.9\% & 31.4\% & 18.7\% & 20.9\% & 32.3\% & 28.1\% & 36.3\% & 37.2\% & 64.1\% & 48.3\% & 46.3\% & 37.8\%\\
\bottomrule
\end{tabular}
}
\end{table*}

\subsection{Implementation Details}
We fine-tune two LLMs on the RbFT task, Llama-3.2-3B-Instruct~\cite{dubey2024llama} and Qwen2.5-3B-Instruct~\cite{yang2024qwen2}, to enhance their robustness against retrieval defects through the LLaMA-Factory toolkit~\footnote{\url{https://github.com/hiyouga/LLaMA-Factory}}. 
In the following text, we refer to these two LLMs as Llama and Qwen for convenience.
The fine-tuning is conducted for 2 epochs, with a learning rate of 1e-5, and a per-device batch size of 16, setting $lora\_rank = 16$ and $lora\_alpha = 64$.  
LLMs are fine-tuned on four types of defective data: \textit{Noisy}, \textit{Irrelevant}, \textit{Counterfactual}, and \textit{Mix}~(randomly selected from the first three types), with the probability of replacing original retrieval results with defective documents~($\tau$) selected from \{0.2, 0.4, 0.6, 0.8, 1.0\}. 
During the evaluation phase, the same $\tau$ values are used, with particular attention given to $\tau=0.4$ and $\tau=1.0$, referred to as the \textit{Normal} and \textit{Hard} settings, respectively, representing moderate and severe retrieval defects.
Additionally, to compare the original performance, we also report their results with the original retrieval results ($\tau = 0$), referred to as the \textit{Clean} setting. 
The retrieval list size $k$ is set to 5. 
Our code and data are available at the URL~\footnote{\url{https://github.com/StibiumT16/Robust-Fine-tuning}}.

\section{Results and Analysis} \label{sec:result}

\begin{figure}[t]
    \centering
    \includegraphics[width=\columnwidth]{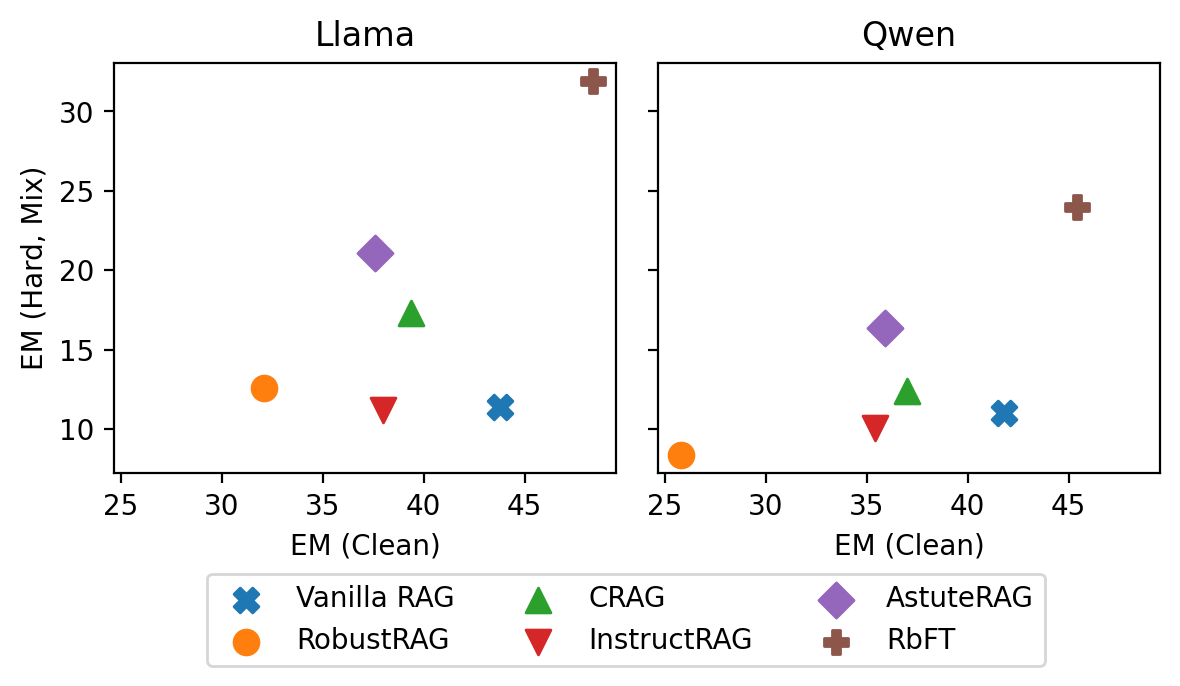}
    \caption{The effectiveness-robustness trade-off scatter diagram. The x-axis represents effectiveness measured by the EM scores of each model in the Clean setting, and the y-axis represents robustness measured by the EM scores of each model in the Hard + Mix setting.}
    \label{fig:tradeoff}
\end{figure}

\begin{figure*}[t]
    \centering
    \includegraphics[width=0.95\textwidth]{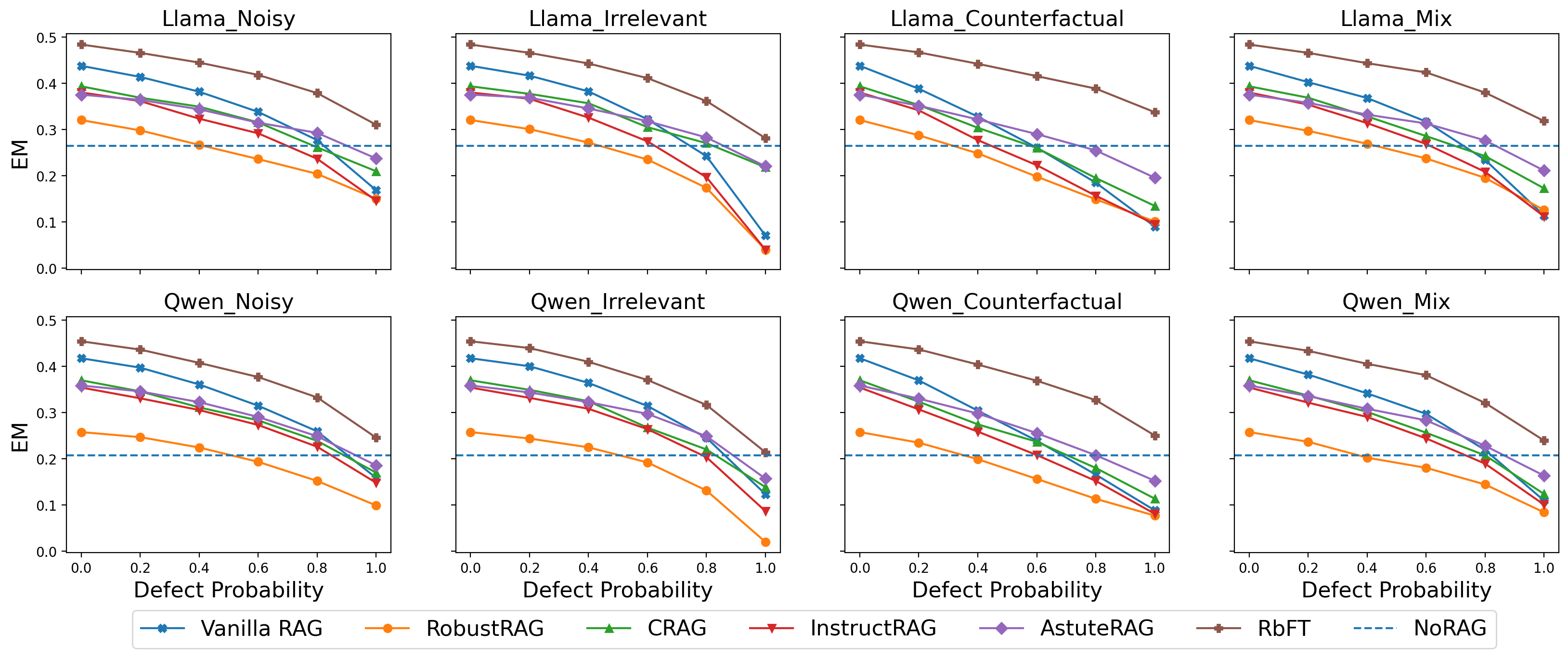}
    \caption{The EM performance of all methods under 4 types of defective data with $\tau = \{0, 0.2, 0.4, 0.6, 0.8, 1.0\}$.}
    \label{fig:mainresult}
\end{figure*}

\subsection{Main Results}
Table~\ref{tab:mainresult} shows the performance of all methods under different levels of retrieval defects. We observe that:

(1) \textbf{In the \textit{Clean} setting, RbFT is the only method that surpasses Vanilla RAG.}
Unlike other approaches that experience performance degradation in defect-free environments, RbFT can enhance robustness while maintaining or even improving performance in such scenarios.
For example, when using Qwen as the base model, RbFT achieves a 12.0\% improvement in F1 score compared to Vanilla RAG.  
This advantage may be attributed to the fact that the top-ranked results returned by the retriever are rarely flawless and may contain a certain proportion of defective documents (i.e., false positives). 
RbFT actively identifies and adapts to these potentially defective documents, thereby preventing performance degradation. 

(2) \textbf{In the \textit{Normal} setting, RbFT consistently achieves the best performance across all retrieval defect scenarios and is still the only method that significantly outperforms Vanilla RAG.}
Notably, RbFT shows the most substantial improvement in the counterfactual defect scenario: when using the Llama model, the EM metric improves by 37.3\% compared to the second-best method, while using the Qwen model also results in a 32.9\% improvement, far exceeding other approaches. 
In other defect scenarios, RbFT also demonstrates strong superiority, with improvements in both metrics mainly ranging between 15\% and 20\%.

\begin{table*}
\caption{Ablation study on the impact of two fine-tuning tasks, Defects Detection (referred to as DD in the table) and Utility Extraction (referred to as UE in the table). The EM metrics on the test set and the change ratios of EM between single-task fine-tuning and RbFT are reported. "*" denotes the result is significantly worse than RbFT with $p<0.05$ level.}
\label{tab:ablation}
\renewcommand{\arraystretch}{1.2}
\scalebox{0.85}{
\begin{tabular}{cc||c|cc|c||c|cc|c}
\toprule
\multicolumn{2}{c||}{LLM} & \multicolumn{4}{c||}{Llama} & \multicolumn{4}{c}{Qwen} \\ 
\hline
\multicolumn{2}{c||}{Method} & Vanilla & Vanilla + DD & Vanilla + UE & RbFT  & Vanilla & Vanilla + DD & Vanilla + UE & RbFT \\ 
\midrule
\multicolumn{2}{c||}{Clean ($\tau=0$)} & 43.8* & 49.7($\uparrow2.7\%$) & 42.7* ($\downarrow11.7\%$) & 48.4 & 41.8* & 45.5($\uparrow0.0\%$) & 41.5* ($\downarrow8.6\%$) & 45.4 \\
\midrule
\multicolumn{1}{c|}{\multirow{4}{*}{\makecell[c]{Normal \\ ($\tau=0.4$)}}} & Noisy & 38.2* & 45.2($\uparrow1.6\%$) & 39.7* ($\downarrow10.8\%$) & 44.5 & 36.1* & 40.7 ($\downarrow0.0\%$) & 37.9* ($\downarrow7.1\%$) & 40.8 \\
\multicolumn{1}{c|}{} & Irrelevant & 38.3* & 43.8($\downarrow1.1\%$) & 39.5* ($\downarrow10.8\%$) & 44.3 & 36.4* & 40.2($\downarrow2.0\%$) & 37.7* ($\downarrow8.0\%$) & 41.0 \\
\multicolumn{1}{c|}{} & Counterfactual & 32.8* & 41.3* ($\downarrow6.6\%$) & 39.9* ($\downarrow9.7\%$) & 44.2 & 30.4* & 37.8*($\downarrow6.4\%$) & 37.1* ($\downarrow8.2\%$) & 40.4 \\
\multicolumn{1}{c|}{} & Mix & 36.8* & 43.7 ($\downarrow1.6\%$) & 39.2 ($\downarrow11.7\%$) & 44.4 & 34.2* & 39.5($\downarrow2.7\%$) & 37.1* ($\downarrow8.6\%$) & 40.6 \\ 
\midrule
\multicolumn{1}{c|}{\multirow{4}{*}{\makecell[c]{Hard \\ ($\tau=1.0$)}}} & Noisy & 16.9* & 28.4*($\downarrow8.7\%$) & 28.9* ($\downarrow7.1\%$) & 31.1 & 15.9* & 22.9* ($\downarrow6.9\%$) & 23.1* ($\downarrow6.1\%$) & 24.6 \\
\multicolumn{1}{c|}{} & Irrelevant & 7.1* & 24.3*($\downarrow13.8\%$) & 25.4* ($\downarrow9.9\%$) & 28.2 & 12.4* & 18.3*($\downarrow14.5\%$) & 18.6* ($\downarrow13.1\%$) & 21.4 \\
\multicolumn{1}{c|}{} & Counterfactual & 9.0* & 22.7*($\downarrow32.8\%$) & 30.3* ($\downarrow10.4\%$) & 33.8 & 8.9* & 16.4*($\downarrow34.7\%$) & 21.6* ($\downarrow13.9\%$) & 25.1 \\
\multicolumn{1}{c|}{} & Mix & 11.4* & 25.7*($\downarrow19.4\%$) & 29.5* ($\downarrow7.5\%$) & 31.9 & 11.0* & 19.5* ($\downarrow18.8\%$) & 21.2* ($\downarrow11.7\%$) & 24.0 \\
\bottomrule
\end{tabular}
}
\end{table*}

(3) \textbf{In the \textit{Hard} setting, RbFT continues to outperform all other methods and further widens the gap with the second-best approach}, highlighting its exceptional performance and adaptability in extremely adverse retrieval environments. 
Traditional methods often exhibit instability when handling these complex issues in such high-difficulty scenarios, where the retrieval results consist entirely of defective documents. 
However, RbFT consistently maintains high performance, particularly in the Counterfactual scenario, where using the Llama model results in an EM metric improvement of over 70\%.
This significant advantage further validates RbFT's robustness and reliability in dealing with various complex retrieval scenarios.

(4) \textbf{RbFT demonstrates significant advantages in balancing effectiveness and robustness.} 
We refer to the EM score under the \textit{Clean} setting as a method’s organic capability in QA tasks (i.e., effectiveness), and refer to the EM score under the \textit{Hard}+ Mix defective setting as the method's ability to handle complex retrieval defects (i.e., robustness). 
Based on these two metrics, we plot the effectiveness-robustness scatter diagram. Results in Figure~\ref{fig:tradeoff} indicate that RbFT outperforms all other methods in both dimensions, achieving the best overall performance. 
Specifically, in terms of effectiveness, RbFT surpasses the second-best method, Vanilla RAG; while in robustness, it significantly outperforms the most competitive method, AstuteRAG. 
This indicates that RbFT not only maintains high answering accuracy but also effectively defends against various complex retrieval defects, striking a better balance between effectiveness and robustness.

\begin{figure*}[t]
    \centering
    \begin{subfigure}{0.85\linewidth}
        \centering
        \includegraphics[width=1\linewidth]{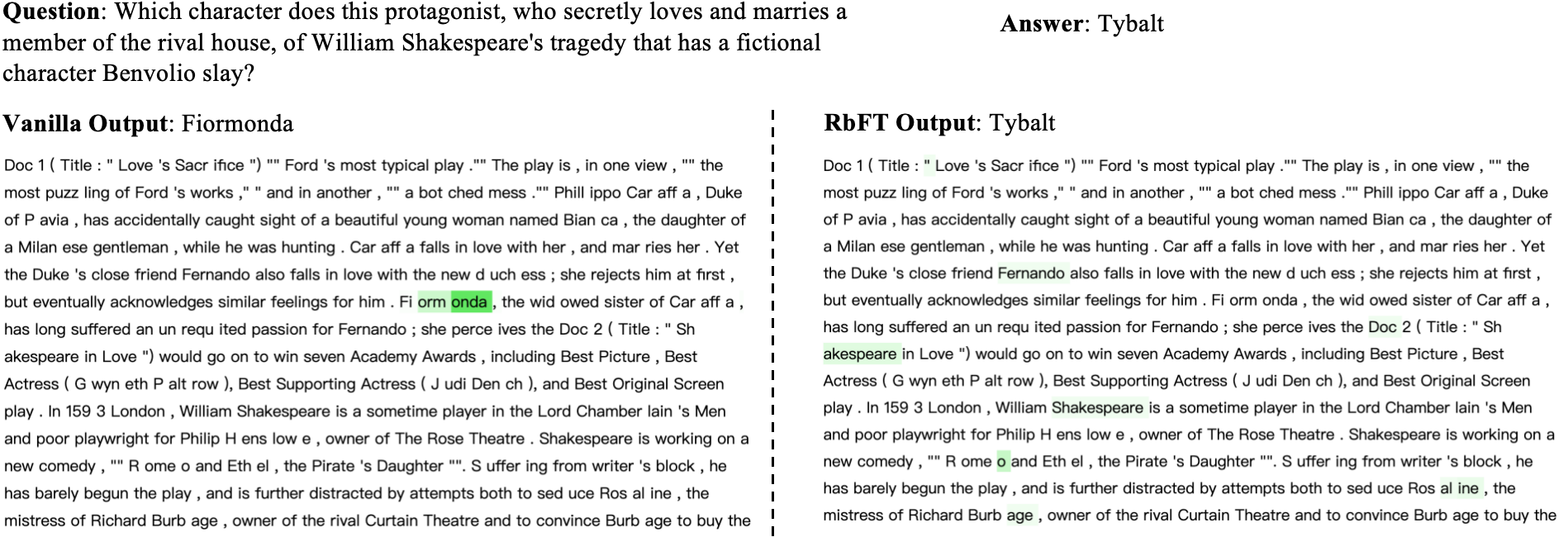}
        \caption{The attention distribution over two input noisy documents of Vanilla RAG and RbFT.}
        \label{sub_fig:nsy}
    \end{subfigure}
    \begin{subfigure}{0.85\linewidth}
        \centering
        \includegraphics[width=1\linewidth]{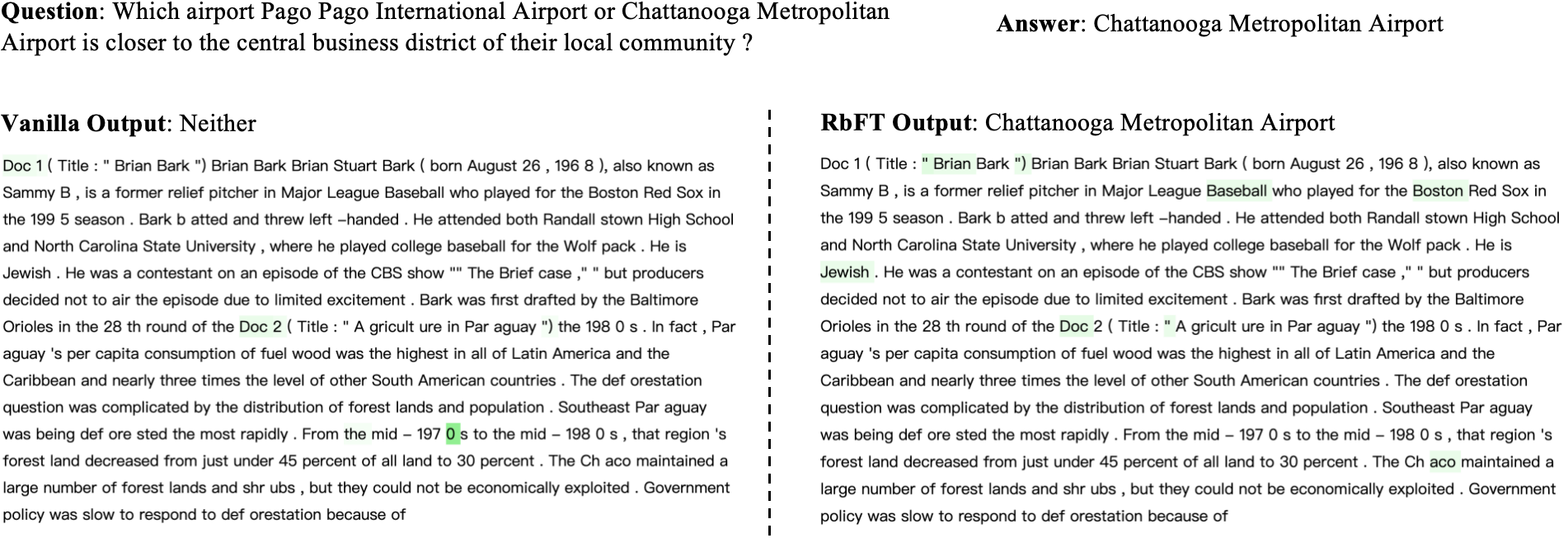}
        \caption{The attention distribution over two input irrelevant documents of Vanilla RAG and RbFT.}
        \label{sub_fig:irr}
    \end{subfigure}
    \begin{subfigure}{0.85\linewidth}
        \centering
        \includegraphics[width=1\linewidth]{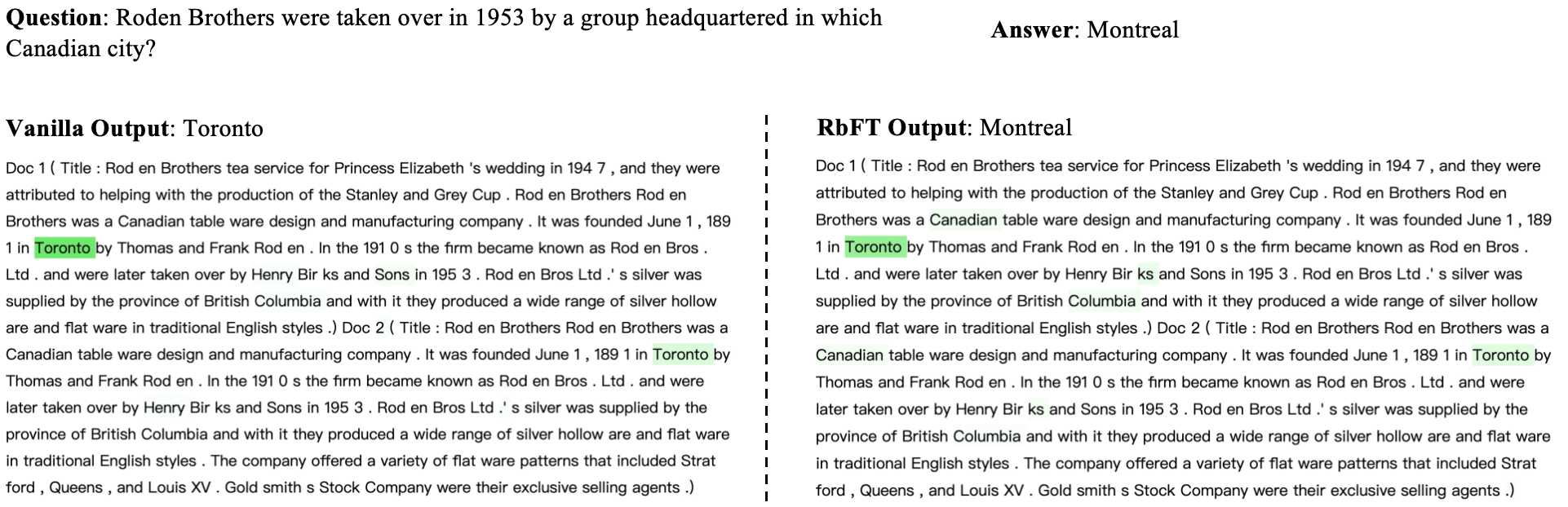}
        \caption{The attention distribution over two input counterfactual documents of Vanilla RAG and RbFT.}
        \label{sub_fig:cf}
    \end{subfigure}
    \caption{Case studies on the attention distribution over input documents of Vanilla RAG and RbFT under different retrieval defects. The greener a document token, the higher the attention it receives during the answer generation process.}
    \label{fig:casestudy}
\end{figure*}

Figure~\ref{fig:mainresult} further illustrates the EM performance of each model under different retrieval defect types and various $\tau$ values. 
It can be observed that RbFT consistently achieves the best performance across all defect levels and retrieval defect scenarios (i.e., Noisy, Irrelevant, Counterfactual, and Mix), demonstrating outstanding robustness and broad applicability. 
Whether in a noise-free standard environment or extreme conditions with highly noisy, irrelevant, or misleading information, RbFT significantly outperforms Vanilla RAG and other enhancement methods. 
Moreover, in high-difficulty scenarios, RbFT further expands its lead over competing approaches, indicating its superior capability in handling harsh retrieval environments.  
This consistent and substantial performance improvement indicates that RbFT is not only highly effective in addressing complex retrieval defects but also better suited for real-world applications, where potential retrieval defects are common. 
As a result, since fluctuations in the quality of retrieved documents are unavoidable in real-world scenarios, our RbFT method, with its capability to provide stable and reliable retrieval-generation responses and maintain strong and consistent robustness, is better suited to meet practical requirements, making it a valuable solution for practical application.

\subsection{Ablation Study}
To further verify the effectiveness and interrelation of the two tasks in RbFT (i.e., Defects Detection and Utility Extraction), we conduct ablation experiments by fine-tuning LLMs using each task individually (referred to as \textit{Vanilla + DD} and \textit{Vanilla + UE}, respectively) to explore their respective roles. 
Specifically, we adopt the same training steps, learning rate, and LoRA parameters as those used in RbFT. 
It is worth noting that, since the instructions and output format of the Defects Detection task differ from those of the original QA tasks, fine-tuning using only Defects Detection data inevitably results in degraded performance on QA tasks.
Therefore, we supplement the training data for \textit{Vanilla + DD} with QA training data in the \textit{Clean} setting (i.e., the original training data).
It can be viewed as the version of RbFT with all defective documents in the Utility Extraction task replaced with their original retrieved versions.

The results of the ablation study, as shown in Table~\ref{tab:ablation}, indicate that both training tasks contribute to improving the robustness of LLMs and RAG systems to some extent, though their improvements are still weaker than RbFT. 
Specifically, the model fine-tuned solely with the Utility Extraction task exhibits a performance drop of approximately 10\% compared to RbFT across all three settings. 
In contrast, the model fine-tuned with Defects Detection demonstrates different features. 
Under settings with weaker retrieval defects (i.e., \textit{Clean} and \textit{Normal}), \textit{Vanilla + DD} achieves performance comparable to RbFT. 
However, in the more challenging retrieval environment of the \textit{Hard} setting, \textit{Vanilla + DD} falls short of \textit{Vanilla + UE} in terms of robustness, especially on the counterfactual data.  
Therefore, the Defects Detection and Utility Extraction training tasks are mutually complementary, working in tandem to reinforce each other's effectiveness.
Only by combining both can we maximize effectiveness in low-defect scenarios while simultaneously enhancing robustness in high-defect environments.

\subsection{Case Study}
In Figure~\ref{fig:casestudy}, we attempt to analyze further how RbFT enhances the defense capability of LLMs by examining the attention distribution over tokens of the input document. 
Specifically, we select one case for each of the three types of retrieval defects (i.e., noisy, irrelevant, and counterfactual) and apply retrieval augmentation to the Llama model using two corresponding defective documents. 
For noisy documents, as shown in Figure~\ref{sub_fig:nsy}, the model fine-tuned with RbFT distributes its attention more evenly across a broader range of contextually relevant information. 
In contrast, the Vanilla model tends to concentrate its attention on distracting and misleading entities, for example, "Fiormonda" in Figure~\ref{sub_fig:nsy}. 
Similarly, in the case of counterfactual documents (as shown in Figure~\ref{sub_fig:cf}), the RbFT-enhanced model focuses less on the incorrect answer "Toronto" and more broadly on multiple relevant pieces of contextual information, thereby mitigating the impact of erroneous and misleading content. 
For irrelevant documents (Figure~\ref{sub_fig:irr}), the Vanilla model also over-focuses on certain specific tokens, whereas the RbFT model distributes its attention more broadly across the context. 
In summary, the attention distribution of LLMs fine-tuned with RbFT becomes smoother compared to the Vanilla LLMs when processing defective input documents. 
This smoother attention distribution helps in two ways. 
First, it increases the model’s resistance against incorrect or irrelevant information by reducing excessive attention to and reliance on such content. 
Second, even when the input document does not directly contain the ground-truth answer, attending to more relevant information in the overall context may better activate the internal parametric knowledge and memory of the LLM, thereby facilitating more accurate responses.

\subsection{Efficiency Analysis}
\begin{table}
\caption{The inference efficiency of each method.}
\label{tab:efficiency}
\renewcommand{\arraystretch}{1.2}
\scalebox{0.9}{
    \begin{tabular}{c|cc}
    \toprule
    \multirow{2}{*}{Method} & \multicolumn{2}{c}{Inference Efficiency (s / query)} \\ 
    \cline{2-3} 
     & Llama & Qwen \\
    \midrule
    Vanilla RAG & 0.193 & 0.198 \\
    RobustRAG & 1.207 & 1.300 \\
    CRAG & 0.401 & 0.401 \\
    InstructRAG & 0.198 & 0.224 \\
    AstuteRAG & 3.417 & 3.369 \\
    RbFT & 0.196 & 0.196 \\
    \bottomrule
    \end{tabular}
}
\end{table}

In Table~\ref{tab:efficiency}, we assess the time efficiency of different methods during inference, reporting the average time required by each RAG system to process a single user query. 
It can be observed that RbFT, by only fine-tuning the LLMs, maintains an inference speed comparable to Vanilla RAG. 
In contrast, other robustness-oriented methods, except for InstructRAG, adopt more complex inference mechanisms that require multiple generation steps or rounds, leading to significantly higher time costs than Vanilla RAG and RbFT.
This demonstrates that RbFT not only excels in performance but also offers a notable advantage in efficiency over other baseline models.
On the other hand, RbFT is vertical to these methods and can be integrated with them to further enhance system robustness.

\section{Conclusion} \label{sec:conclu}
In this work, we introduce Robust Fine-Tuning (RbFT), a novel fine-tuning approach to enhance the robustness of RAG systems against retrieval defects. 
By addressing the critical vulnerabilities in RAG systems, specifically their susceptibility to defective retrieval results, RbFT equips LLMs with improved defensive capabilities. 
Our dual-task fine-tuning strategy mitigates the impact of defective retrieval inputs and ensures effective knowledge utilization even under adverse retrieval conditions.
Extensive experimental evaluations demonstrate that RbFT significantly outperforms existing state-of-the-art methods in terms of robustness and inference efficiency. 
Notably, RbFT maintains high effectiveness even in clean environments while offering reliable responses in high-defect settings, making it a robust and practical solution for real-world RAG applications.
In future works, we plan to extend RbFT beyond QA tasks to a broader range of applications. 
Additionally, since RbFT is theoretically vertical to other baselines focusing on inference strategies and mechanisms, we intend to explore their integration to develop more efficient and robust RAG systems further.

\bibliographystyle{ACM-Reference-Format}
\bibliography{sample-base}

\end{document}